\documentclass{article}
\usepackage{ijcai26}

\usepackage{times}
\usepackage{soul}
\usepackage{url}
\usepackage[hidelinks]{hyperref}
\usepackage[utf8]{inputenc}
\usepackage[small]{caption}
\usepackage{graphicx}
\usepackage{amsmath}
\usepackage{amsthm}
\usepackage{amssymb}
\usepackage{booktabs}
\usepackage{algorithm}
\usepackage{algorithmic}
\usepackage[switch]{lineno}

\usepackage{comment}
\usepackage{xcolor}
\usepackage{colortbl}
\usepackage{multirow}
\usepackage{booktabs}
\usepackage{tabularx}
\usepackage{array}
\newcolumntype{C}[1]{>{\centering\arraybackslash}p{#1}}

\title{Physics-Guided Geometric Diffusion for Macro Placement Generation}

\author{
Jongho Yoon$^{1}$   % \footnote{Both authors contributed equally to this paper.}
\and
Jinsung Jeon$^{2,3} $  % \footnote{Work done while the author was at UC San Diego.}
\and
Seokhyeong Kang$^4$\\
\affiliations
$^1$POSTECH Institute of Artificial Intelligence\\
$^2$KAIST InnoCORE LLM \\
$^3$Seoul National University \\
$^4$Pohang University of Science and Technology\\
\emails
\{jh.yoon, shkang\}@postech.ac.kr,
jjsjjs0902@gmail.com
}
\begin{document}

\maketitle

\begin{abstract}
Macro placement is a pivotal stage in VLSI physical design, fundamentally determining the overall chip performance.
Recent data-driven placement methods have demonstrated significant potential, yet they often struggle to handle sequential dependencies and to balance topological connectivity with physical constraints. 
To bridge this gap, we propose \textbf{MacroDiff+}, a physics-guided geometric diffusion framework.
Specifically, we design a dual-domain denoising architecture that couples topological connectivity encoded by heterogeneous GNNs with global geometric context modeled by a Transformer. 
Furthermore, we introduce Physics-Guided Sampling, an inference strategy that actively steers the generation using explicit gradients to ensure both statistical plausibility and physical validity.
On the ISPD2005 MMS benchmarks, MacroDiff+ outperforms state-of-the-art baselines with a 6.1–6.2\% reduction in wirelength. Notably, it exhibits superior stability and scalability on large-scale designs where prior methods fail to converge.
The source code is provided at \url{https://github.com/jhy00n/MacroDiff-plus}.
\end{abstract}

\section{Introduction}
Macro placement is the stage of physical design that determines the locations of large, pre-designed circuit blocks, such as memory arrays, IP cores, and analog blocks, on the chip canvas.
As modern system-on-chip (SoC) designs continue to scale in complexity, incorporating a growing number of macros alongside numerous standard cells, the quality of macro placement has become a pivotal determinant of chip success~\cite{agnesina2023autodmp}. 
Decisions made at this stage create cascading effects throughout subsequent design phases, ultimately determining the power, performance, and area (PPA) metrics~\cite{tseng2024challenges}. 

However, macro placement is an inherently challenging combinatorial optimization problem.
The search space grows exponentially with the number of macros, as each macro must be assigned both a location and an orientation on the chip canvas~\cite{mirhoseini2021graph}.
This complexity makes exhaustive exploration computationally intractable, and manual placement by experienced engineers requires weeks or even months of iterative effort, creating a critical bottleneck in the design cycle~\cite{gao2022congestion}. 
These challenges have driven the development of automated macro placement methods to efficiently explore the design space and deliver consistent, high-quality solutions.

To address these complexities, automated macro placement has primarily followed two paradigms: optimization-based and reinforcement learning (RL)-based approaches.
Optimization-based methods~\cite{agnesina2023autodmp,shi2023macro} formulate placement as a continuous or black-box optimization problem, leveraging powerful solvers to navigate the design space. 
Similarly, RL-based methods~\cite{mirhoseini2021graph,lai2022maskplace,lai2023chipformer} frame the task as a sequential Markov Decision Process, placing macros one at a time.
However, despite their advancements, these approaches face fundamental limitations from an AI methodology perspective. 
Optimization solvers often struggle with the high dimensionality and become trapped in local optima, while the sequential formulation of RL introduces inherent order dependency, preventing the simultaneous optimization of global interdependencies.

Recently, diffusion models have emerged as a promising generative paradigm that addresses some of these limitations~\cite{ho2020denoising,trippe2023diffusion,sun2023difusco}. 
By generating all macro positions simultaneously through iterative denoising, diffusion models inherently capture global context without the order dependency of RL, while their stochastic sampling naturally explores diverse solutions beyond the local optima that trap optimization-based solvers.
However, existing diffusion-based approaches for placement~\cite{lee2025chip} treat the problem purely as geometric pattern generation, failing to explicitly model the circuit netlist topology that fundamentally governs wirelength optimization. 
This disconnect between geometric generation and topological objectives leads to unstable convergence and suboptimal wirelength, as evidenced by the high variance in the prior work.

To reconcile this disconnect, we propose \textbf{MacroDiff+}, a physics-guided geometric diffusion framework that explicitly synergizes topological connectivity with spatial constraints.
The main contributions are as follows:
\begin{itemize}
\item Dual-Branch Denoising Architecture: We propose a novel architecture that integrates Heterogeneous Graph Neural Networks (Hetero GNN) for topological encoding and Transformers for global spatial context.
\item Physics-Guided Sampling: We introduce an inference-time mechanism that steers the generative trajectory using explicit gradients of physical objectives, ensuring generated solutions satisfy hard constraints beyond what the learned distribution captures.
\item Empirical Validation: Experiments on ISPD2005 benchmarks demonstrate 6.1–6.2\% HPWL reduction over RL-based and optimization-based baselines, significantly improved stability, and successful scaling to large designs where prior methods fail to converge.
\end{itemize}

\section{Background and Related Work}

\subsection{Macro Placement in VLSI}
Macro placement significantly impacts the PPA metrics of the final chip layout. 
Since poor placement decisions can lead to irreversible routing failures and timing violations, achieving an optimal macro configuration is critical~\cite{chen2023macrorank,xue2024reinforcement}. 
Classical heuristics such as Simulated Annealing~\cite{sechen2003timberwolf} and Force-Directed methods~\cite{chan2005multilevel} established foundational concepts in this domain. However, these methods struggle with the scalability and heterogeneity of modern SoC designs~\cite{qiu2023progress}. 
Consequently, recent research has shifted towards advanced optimization and machine learning (ML) techniques to navigate this complex combinatorial search space.

\subsubsection{Optimization-based Approaches} 
Optimization-based methods typically formulate placement as a continuous or black-box optimization problem. 
For instance, AutoDMP~\cite{agnesina2023autodmp} enhances mixed-size placement quality through the automated parameter tuning of the analytical placer DREAMPlace~\cite{chen2023stronger}. It utilizes a Multi-Objective Tree-structured Parzen Estimator (Bayesian optimization) to efficiently optimize PPA proxy objectives such as wirelength and congestion. Similarly, WireMask-BBO~\cite{shi2023macro} approaches macro placement as a black-box optimization problem. Instead of purely analytical solvers, it represents solutions via continuous coordinates and employs a wire-mask-guided evaluation to greedily improve placement using evolutionary algorithms.

While these approaches leverage powerful solvers to minimize objectives like Half-Perimeter Wirelength (HPWL) and demonstrate effectiveness in local refinement, they face significant challenges. They often suffer from excessive runtime, struggle to navigate the non-convex optimization landscape characterized by discrete physical constraints, and their performance heavily relies on the quality of initialization.

\subsubsection{RL-based Approaches} 
Leveraging advancements in deep learning, RL-based methods typically frame macro placement as a sequential Markov Decision Process (MDP).
Chip Placement with Deep RL~\cite{mirhoseini2021graph} garnered significant attention by demonstrating that an RL agent could outperform human experts in hours by predicting optimal positions for cells sequentially. 
Building on this, MaskPlace~\cite{lai2022maskplace} recasts the problem as learning pixel-level visual representations. It utilizes a convolutional neural network to process position, wire, and view masks, and employs a dense reward based on incremental HPWL changes to guarantee non-overlapping placements. 
Furthermore, ChiPFormer~\cite{lai2023chipformer} advances this domain by utilizing a Decision Transformer for offline RL, enabling transferable policies that generalize to new circuits with minimal fine-tuning.

Despite their potential to learn complex policies, these RL-based approaches face inherent structural limitations. 
The sequential nature of placement decisions introduces order dependency, where early suboptimal decisions propagate, making it difficult to optimize the global context of all macros simultaneously. 
Moreover, these methods are often plagued by significant sample inefficiency and incur substantial computational costs or extensive retraining on unseen designs.

\subsection{Generative Diffusion Models}
\label{sec:diffusion_background}

Diffusion models have emerged as a powerful generative paradigm, capable of modeling complex continuous distributions by learning to reverse a gradual noise addition process~\cite{ho2020denoising}. Unlike RL-based methods that rely on sequential decision-making, diffusion models generate holistic solutions for all objects simultaneously, making them well-suited for the combinatorial nature of chip placement.

\subsubsection{Fundamentals of Diffusion Models}
As a representative framework of diffusion models, Denoising Diffusion Probabilistic Models (DDPMs)~\cite{ho2020denoising} characterize the generation process through two Markov chains: a fixed \textit{forward process} ($q$) that gradually adds Gaussian noise, and a learned \textit{reverse process} ($p_\theta$) that iteratively recovers the original data.
The core objective is to train a network $\epsilon_\theta(x_t, t)$ to predict the noise $\epsilon$ added to the input.
This is achieved by minimizing the simplified mean squared error (MSE):
\begin{equation}
\label{eq:loss}
    \mathcal{L}_{\text{simple}}(\theta) = \mathbb{E}_{t, x_0, \epsilon} \left[ \| \epsilon - \epsilon_\theta(x_t, t) \|^2 \right].
\end{equation}
By minimizing this objective, the model implicitly learns the score function $\nabla_x \log p(x)$, which guides the generation from random noise to a structured layout. We provide the detailed mathematical preliminaries in Appendix~\ref{app:diffusion}.

% Formally, a Denoising Diffusion Probabilistic Model (DDPM) involves two processes: a fixed \textit{forward process} ($q$) that gradually adds Gaussian noise to data $x_0$ over $T$ steps, and a learned \textit{reverse process} ($p_\theta$) that iteratively denoises $x_t$ to recover the original distribution. While detailed formulations of the Markov chains are provided in Appendix~\ref{app:diffusion}, the core objective is to train a network $\epsilon_\theta(x_t, t)$ to predict the noise $\epsilon$ added to the input. This is achieved by minimizing the simplified mean squared error (MSE) loss:
% \begin{equation}
% \label{eq:loss}
%     \mathcal{L}_{\text{simple}}(\theta) = \mathbb{E}_{t, x_0, \epsilon} \left[ \| \epsilon - \epsilon_\theta(x_t, t) \|^2 \right].
% \end{equation}
% This objective effectively allows the model to learn the score function $\nabla_x \log p(x)$, guiding the generation from random noise to a structured layout.

\subsubsection{Generative Diffusion for Macro Placement}
Recently, ChipDiffusion~\cite{lee2025chip} pioneered the application of this paradigm to VLSI, framing macro placement as an image generation task. By leveraging the continuous spatial modeling capabilities of diffusion, it overcomes the local optima issues of optimization-based methods and the sequential limitations of RL. 
However, existing diffusion approaches typically rely on image-based representations or synthetic data, frequently overlooking the precise \textit{topological connectivity} inherent in real-world netlists. By treating placement purely as a geometric generation task, they lack explicit mechanisms to balance wirelength minimization with strict physical constraints (e.g., overlaps) during denoising.

\begin{figure*}[t]
  \centering
  \includegraphics[width=177.6mm]{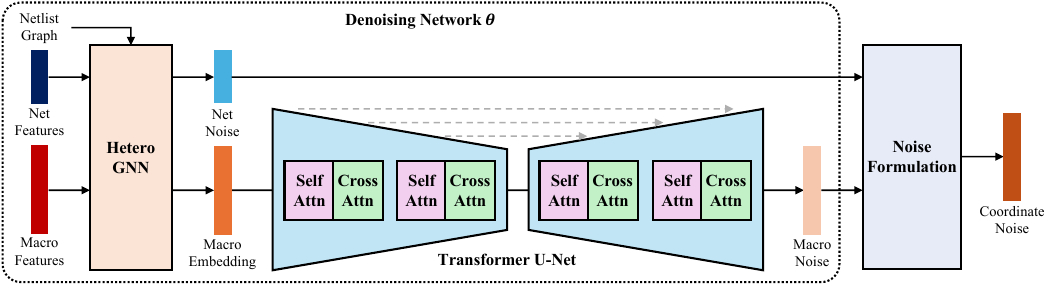}
  \caption{Overall framework of the dual-branch noise prediction network in \textbf{MacroDiff+}. The architecture integrates topological insights from the Hetero GNN, yielding Macro Embeddings ($H_{macro}$) and Net-Centric Noise ($\epsilon_{net}$), with global geometric context from the Transformer producing Macro-Centric Noise ($\epsilon_{cell}$). The Noise Formulation module fuses these outputs to generate the final Coordinate Noise, $\epsilon_\theta(x_t, t)$.}
  \label{fig:architecture}
\end{figure*}

\section{Proposed Method} \label{sec:method}
To overcome these limitations, we propose \textbf{MacroDiff+}, a physics-guided graph diffusion framework.
Unlike prior geometric-only approaches, our method explicitly models the circuit topology using a heterogeneous graph and steers the generation process based on physical gradients.

\subsection{Problem Formulation and Training Objective}
\label{sec:formulation}
We formulate macro placement as a generative task on a heterogeneous graph $\mathcal{G} = (\mathcal{V}, \mathcal{E})$.
The node set $\mathcal{V}$ is partitioned into macros $\mathcal{V}_{m}$ and nets $\mathcal{V}_{n}$, where edges $\mathcal{E}$ represent pin-level connectivity.
Each macro node $v_i \in \mathcal{V}_{m}$ is characterized by its fixed dimensions $(w_i, h_i)$, while the placement solution is defined by the coordinates $x \in \mathbb{R}^{|\mathcal{V}_{m}| \times 2}$ representing the bottom-left corners.
Our objective is to generate the optimal positions $x$ that minimize the Half-Perimeter Wirelength (HPWL) while strictly satisfying non-overlap constraints. 
Our heterogeneous graph encodes both static circuit properties (macro dimensions, chip boundaries, net degrees, pin offsets) and dynamic placement states (macro positions $x_t$ and net-level HPWL) that evolve during denoising. 
The pin-offset feature on edges enables wirelength computation at the pin level, which is critical for large macros.
Detailed specifications are provided in Appendix~\ref{app:features}.

To train our denoising network $\theta$, we follow the standard diffusion training paradigm. Specifically, the model is optimized to predict the noise $\epsilon$ added to the clean placement $x_0$ at a given timestep t. The Training objective is defined by minimizing the mean squared error between the added noise and the noise predicted by our dual-branch architecture, as formulated in Eq.~\ref{eq:loss}. By optimizing this objective, the model learns to capture the joint distribution of circuit connectivity and spatial constraints, enabling the generation of high-quality macro placements from random noise.

\subsection{Dual-Branch Denoising Architecture}
\label{sec:architecture}
To generate high-quality placements, the model must simultaneously capture local connectivity (topology) and global spatial relationships (geometry). 
We design a dual-branch architecture (cf. Fig.~\ref{fig:architecture}) that handles these two modalities via a Hetero GNN and a Transformer, respectively.

\subsubsection{Heterogeneous Graph Neural Network (Hetero GNN)} \label{method-a}
The Hetero GNN serves as the topological backbone of our framework, specifically designed to handle the non-Euclidean structure of the netlist.
Since macro placement is fundamentally driven by connectivity, where macros linked by critical nets must be placed in proximity, capturing these dependencies is crucial.
We employ a relational graph attention to explicitly model these interactions between macros and nets.

Conditioned on timestep $t$, the Hetero GNN iteratively updates node embeddings via message passing. 
Formally, for each layer $l$, the update process for a macro node $m$ aggregating messages from neighboring net nodes $n \in \mathcal{N}(m)$ is defined as:
\begin{equation}
    \mathbf{e}^{(l+1)}_{n\rightarrow m} = \text{AGGREGATE}^{(l)}_{n\rightarrow m}(\{h_n^{(l)} : n \in \mathcal{N}(m)\}, t),
\end{equation}
\begin{equation}
    h_m^{(l+1)} = \text{UPDATE}^{(l+1)}(h_m^{(l)}, \mathbf{e}^{(l+1)}_{n\rightarrow m}).
\end{equation}
In our implementation, the $\text{AGGREGATE}$ function employs the GATv2 attention mechanism~\cite{brody2021attentive}.
This allows the model to dynamically weigh the importance of information from different neighboring nodes based on their features, enabling a more nuanced understanding of connectivity than methods with static weights.
The $\text{UPDATE}$ function then combines the resulting aggregated message with the previous node embedding $h_m^{(l)}$ through a non-linear transformation and a residual connection.
The residual connection provides a direct path for gradients, mitigating the vanishing gradient problem in deep Hetero GNNs.
A symmetric process is applied to update net embeddings ($h_n$) by aggregating information from macro nodes.
Through $L$ layers of such message passing, the model learns the structural importance of each net and macro, producing topologically rich embeddings $H_{macro}$ and $H_{net}$.

The net embeddings $H_{net}$ serve as the Net-Centric Noise ($\epsilon_{net}$), a high-dimensional vector defined on net nodes that encodes the connection intent, the learned importance of each net at the current denoising step.
Conceptually, this allows the model to directly incorporate the critical HPWL objective into noise prediction, identifying which connections are most critical to optimize.
However, a key distinction is that despite being termed noise by analogy with the standard diffusion formulation, $\epsilon_{net}$ resides in an abstract topological space rather than the physical coordinate space, representing connection intent rather than spatial displacement. 
To convert this intent into geometric updates, our fusion phase introduces a projection mechanism, described next.

\subsubsection{Transformer Network} % \label
Complementing the Hetero GNN, the Transformer network serves as the geometric branch.
It is built upon a U-Net architecture integrated with Transformer blocks.
The hierarchical structure of the U-Net processes spatial features at multiple scales, while the embedded Transformer blocks perform two critical attention operations to resolve geometric constraints.
First, \textit{Self-attention} captures long-range dependencies among all macros, enabling the model to detect and diffuse high-density clusters effectively.
Second, \textit{Cross-attention} injects physical conditioning information $C$ into the latent space. Here, $C$ corresponds to the macro size features $(w, h)$ (see Appendix~\ref{app:features} for detailed input features), ensuring the generation process remains cognizant of physical dimensions.
Specifically, the macro embeddings $H_{macro}$ from the Hetero GNN form the initial sequence $Z^{(0)}$.
Formally, the operation sequence within each Transformer block at layer $l$ is defined as:
\begin{equation}
    Z'^{(l)} = \text{Self-Attn}(Z^{(l-1)}, t),
\end{equation}
\begin{equation}
    Z''^{(l)} = \text{Cross-Attn}(Z'^{(l)}, C),
\end{equation}
\begin{equation}
    Z^{(l)} = \text{FFN}(Z''^{(l)}).
\end{equation}
Here, each function includes residual connections and normalization.
Finally, a projection head converts the output $Z^{(L)}$ into the Macro-Centric Noise ($\epsilon_{cell}$).

Unlike the net-centric prediction from the GNN branch, this macro-centric prediction is a direct coordinate-based noise prediction inferred from the global spatial context.
Its primary strength lies in generating well-distributed layouts that respect chip boundaries and density constraints.
However, a significant limitation is that this prediction is agnostic to the netlist topology; it may produce valid but wirelength-suboptimal placements. 
Therefore, it serves as the geometric counterpart to the topological insight provided by the Hetero GNN, necessitating the synergistic fusion mechanism we detail in the next section.

\subsubsection{Dual-Branch Fusion and Noise Prediction}\label{method-c}
\begin{figure}[t!]
\centering
\begin{center}
\includegraphics[width=85.1mm]{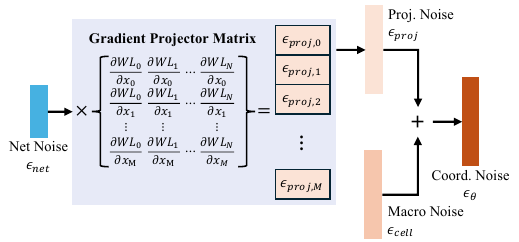}
\caption{Detailed schematic of the Noise Formulation.}
\label{fig:noiseprediction}
\end{center}
\end{figure}

To harness the strengths of both branches, we formulate the final noise prediction as a synergistic fusion within the Noise Formulation module.
This fusion utilizes the \textbf{Gradient Projector Matrix} as a bridge between the topological and geometric domains (cf. Fig.~\ref{fig:noiseprediction}).
Since the net-centric noise $\epsilon_{net}$ resides in an abstract topological space, it cannot directly update macro coordinates.
To resolve this, we employ a projection mechanism using the gradient of the HPWL. Specifically, we project $\epsilon_{net}$ into the coordinate space via the Gradient Projector Matrix (implemented as a Jacobian vector-product) and combine it with the macro-centric noise $\epsilon_{cell}$ as follows:
\begin{equation}
\label{eq:noise}
\epsilon_\theta(x_t,t) = \lambda_{cell} \cdot \epsilon_{cell} + \lambda_{net} \cdot (\nabla_{x_t}\text{HPWL}(x_t))^T \cdot \epsilon_{net}.
\end{equation}
Here, the term $(\nabla_{x_t}\text{HPWL}(x_t))^T$ corresponds to the Gradient Projector Matrix shown in Fig.~\ref{fig:noiseprediction}. It represents the sensitivity of wirelength to macro displacements, effectively translating the learned connection importance into physical force vectors. The scalars $\lambda_{cell}$ and $\lambda_{net}$ are hyperparameters that balance the influence of the two branches. This formulation (Eq.~\ref{eq:noise}) mathematically realizes the synergy proposed in our framework, balancing global spatial arrangement (density) with fine-grained connectivity objectives (wirelength).

\begin{figure*}[t!]\centering
\begin{center}
\includegraphics[width=178.1mm]{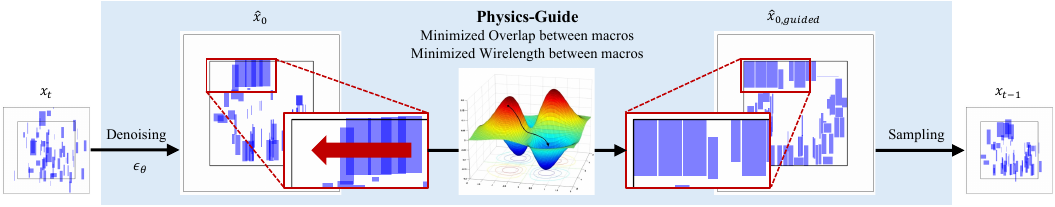}
\caption{Detailed workflow of the Physics-Guided Sampling employed during inference.}
\label{fig:Physics_guided}
\end{center}
\end{figure*}

\subsection{Physics-Guided Sampling}
\label{sec:sampling}
While the dual-branch model learns the distribution of high-quality placements, probabilistic generation alone may yield samples violating strict physical constraints. 
We thus employ Physics-Guided Sampling (cf. Fig.~\ref{fig:Physics_guided}), which refines each predicted placement to better satisfy physical constraints during generation.

The procedure operates by augmenting the standard reverse diffusion process. 
At each timestep $t$, we first use our trained dual-branch denoising model $\epsilon_\theta$ to obtain an initial prediction of the clean macro placement, denoted as $\hat{x}_0$.
This prediction is derived from the current noisy state $x_t$ using the standard diffusion formulation:
\begin{equation}
\label{eq:predict0}
    \hat{x}_0 = \frac{1}{\sqrt{\bar{\alpha}_t}}(x_t - \sqrt{1 - \bar{\alpha}_t}\epsilon_\theta(x_t,t)).
\end{equation}

This initially predicted placement $\hat{x}_0$ serves as the starting point for a brief, physics-driven optimization. 
We define a composite cost function, $\mathcal{L}_{guide}$, that mathematically represents the quality of the placement:
\begin{equation}
\label{eq:guide}
\begin{split}
    \mathcal{L}_{guide}(\hat{x}_0) = & w_{HPWL} \cdot \mathcal{L}_{HPWL}(\hat{x}_0) \\
    & + w_{overlap} \cdot \mathcal{L}_{overlap}(\hat{x}_0),
\end{split}
\end{equation}
where $\mathcal{L}_{HPWL}$ estimates the wirelength and $\mathcal{L}_{overlap}$ penalizes illegal overlaps. 
The scalars $w_{HPWL}$ and $w_{overlap}$ follow a two-phase schedule: the first phase emphasizes wirelength minimization, and once HPWL improvement plateaus, the second phase down-weights $L_{HPWL}$ to focus on overlap resolution. 
This scheduling avoids gradient interference between conflicting objectives that arises under fixed weights.
To ensure the guidance is physically meaningful, we perform this optimization directly in the clean data space ($\hat{x}_0$) rather than the noisy latent space. 
The goal is to find a modification vector, $\Delta{x_0}$, that minimizes the cost:
\begin{equation}
    \min_{\Delta}{\mathcal{L}_{guide}(\hat{x}_0+\Delta)}.
\end{equation}
In practice, we approximate this minimization via gradient descent steps starting from $\Delta = 0$. This step effectively pulls the predicted placement towards a configuration that simultaneously reduces wirelength and resolves overlaps. The refined placement is obtained as:
\begin{equation}
\hat{x}_{0,guided} = \hat{x}_0 + \Delta x_0.
\label{eq:guided_0}
\end{equation}
Crucially, this refined placement $\hat{x}_{0,guided}$ is not the final output but is used to guide the sampling trajectory. We derive an implicitly guided noise, $\epsilon_{guided}$, based on this physically optimized prediction:
\begin{equation}
\epsilon_{guided} = \frac{1}{\sqrt{1-\bar{\alpha}_t}}(x_t - \sqrt{\bar{\alpha}_t}\hat{x}_{0,guided}).
\label{eq:denoiseonestep}
\end{equation}
By replacing the original network prediction $\epsilon_\theta$ with $\epsilon_{guided}$ in the update rule for $x_{t-1}$, we ensure that every step of the generation is not only grounded in the learned data distribution but is also actively steered by explicit physical forces. This procedure, outlined in Algorithm~\ref{alg:sample}, allows MacroDiff+ to produce layouts that are both topologically coherent and physically valid.

\begin{algorithm}[t]
\caption{Physics-Guided Sampling from MacroDiff+}
\label{alg:sample}
\begin{algorithmic}[1]
\REQUIRE Learned model $\epsilon_{\theta}$, graph $G$ with static features, step size $\eta$, guidance steps $K$, weights $w_{\text{HPWL}}, w_{\text{overlap}}$, weight schedule $W(\cdot)$
\STATE $x_T \sim \mathcal{N}(0, I)$
\STATE Construct graph $G_T$ using $x_T$
\FOR{$t=T, \dots, 1$}
    \STATE $z \sim \mathcal{N}(0, I)$ if $t>1$, else $z=0$
    \STATE $\epsilon_{\text{pred}} \leftarrow \epsilon_\theta(G_t, t)$ \COMMENT{Predict noise}

    \STATE \textit{// Step 1: Estimate clean data (Eq.~\ref{eq:predict0})}
    \STATE $\hat{x}_0 \leftarrow \frac{1}{\sqrt{\bar{\alpha}_t}} \left( x_t - \sqrt{1 - \bar{\alpha}_t}\epsilon_{\text{pred}} \right)$

    \STATE \textit{// Step 2: Physics-Guided Optimization (Eq. \ref{eq:guide}-\ref{eq:guided_0})}
    \STATE $\hat{x}_{0, \text{guided}} \leftarrow \hat{x}_0$
    \FOR{$k=1, \dots, K$}
        \STATE $(w_{HPWL}, w_{overlap}) \leftarrow W(k, L_{HPWL})$
        \STATE $L_{\text{guide}} \leftarrow w_{\text{HPWL}} \cdot L_{\text{HPWL}} + w_{\text{overlap}} \cdot L_{\text{overlap}}$
        \STATE $\nabla \leftarrow \nabla_{\hat{x}_{0, \text{guided}}} L_{\text{guide}}$ \COMMENT{Compute gradient}
        \STATE $\hat{x}_{0, \text{guided}} \leftarrow \hat{x}_{0, \text{guided}} - \eta \cdot \nabla$ \COMMENT{Gradient descent}
    \ENDFOR

    \STATE \textit{// Step 3: Compute guided noise (Eq.~\ref{eq:denoiseonestep})}
    \STATE $\epsilon_{\text{guided}} \leftarrow \frac{1}{\sqrt{1 - \bar{\alpha}_t}} \left( x_t - \sqrt{\bar{\alpha}_t} \hat{x}_{0, \text{guided}} \right)$

    \STATE \textit{// Step 4: Denoise one step}
    \STATE $x_{t-1} \leftarrow \frac{1}{\sqrt{\alpha_t}} \left( x_t - \frac{1-\alpha_t}{\sqrt{1-\bar{\alpha}_t}} \epsilon_{\text{guided}} \right) + \sigma_t z$
    \STATE Construct graph $G_{t-1}$ using $x_{t-1}$
\ENDFOR
\RETURN $x_0$
\end{algorithmic}
\end{algorithm}

\begin{comment}
\end{comment}
\section{Experimental Evaluations}

\begin{table*}[t]
\centering
\footnotesize
\newcolumntype{Y}{>{\centering\arraybackslash}X}
\begin{tabularx}{\textwidth}{l c c@{\,$\pm$\,}c Y c@{\,$\pm$\,}c Y c@{\,$\pm$\,}c Y c@{\,$\pm$\,}c Y}
\toprule
\multirow{2.5}{*}{Design} & \multirow{2.5}{*}{DREAMPlace} & \multicolumn{3}{c}{MaskPlace} & \multicolumn{3}{c}{WireMask} & \multicolumn{3}{c}{ChipDiffusion} & \multicolumn{3}{c}{MacroDiff+ (Ours)} \\
\cmidrule(lr){3-5} \cmidrule(lr){6-8} \cmidrule(lr){9-11} \cmidrule(lr){12-14}
& & \multicolumn{2}{c}{Mean $\pm$ Std} & Best & \multicolumn{2}{c}{Mean $\pm$ Std} & Best & \multicolumn{2}{c}{Mean $\pm$ Std} & Best & \multicolumn{2}{c}{Mean $\pm$ Std} & Best \\
\midrule
\textit{adaptec1} & 64.7  & 72.1 & 1.5 & 69.3 & 71.0 & 2.1 & \textbf{68.0} & 70.1 & 1.1 & 68.3 & \textbf{69.1} & \textbf{1.0} & 68.1 \\
\textit{adaptec2} & 75.8  & 109.0 & 12.3 & 90.8 & 108.8 & 8.1 & 95.6 & 96.0 & 20.9 & 80.5 & \textbf{81.6} & \textbf{2.6} & \textbf{79.4} \\
\textit{adaptec3} & 153.3 & 185.3 & 5.8 & 178.3 & 180.8 & 4.8 & 173.3 & 165.2 & 1.3 & 164.0 & \textbf{163.9} & \textbf{1.1} & \textbf{162.4} \\
\textit{adaptec4} & 142.4 & 162.5 & 1.9 & 160.6 & 164.7 & 4.4 & 159.0 & 150.6 & 1.2 & 149.0 & \textbf{147.9} & \textbf{0.6} & \textbf{146.9} \\
\textit{bigblue1} & 85.3  & 89.0 & 1.7 & 87.3 & 88.2 & 0.6 & 87.2 & \textbf{87.2} & \textbf{0.5} & 86.8 & 87.3 & 0.5 & \textbf{86.7} \\
\textit{bigblue2} & 125.4 & \multicolumn{2}{c}{T/O} & T/O & \multicolumn{2}{c}{T/O} & T/O & \textbf{134.9} & \textbf{0.5} & \textbf{134.4} & 139.5 & 1.0 & 138.1 \\
\textit{bigblue3} & 273.8 & 325.7 & 13.7 & 307.1 & 327.1 & 17.4 & 313.2 & 319.6 & 5.0 & 310.3 & \textbf{299.3} & \textbf{4.3} & \textbf{293.8} \\
\textit{bigblue4} & 643.2 & \multicolumn{2}{c}{T/O} & T/O & \multicolumn{2}{c}{T/O} & T/O & 702.8 & 10.4 & 688.8 & \textbf{678.9} & \textbf{7.0} & \textbf{672.0} \\
\bottomrule
\end{tabularx}
\caption{Comparison of mixed-size placement results on HPWL ($\times 10^6$). Each method performs macro placement using its respective approach, followed by standard cell placement using DREAMPlace. Lower values are better.}
\label{table:results}
\end{table*}

\begin{figure*}[t]\centering
\begin{center}
\includegraphics[width=178.1mm]{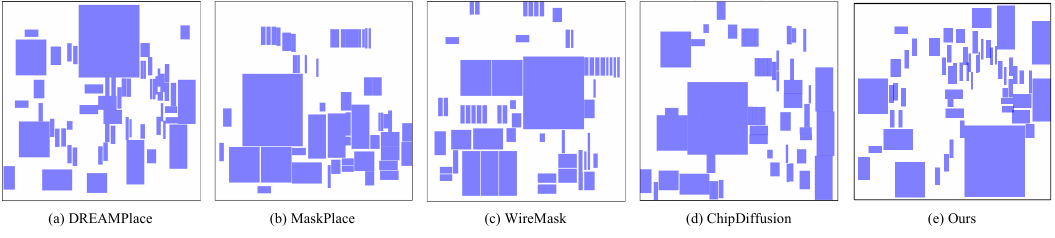}
\caption{Qualitative comparison on adaptec3. (e) MacroDiff+ achieves a more uniform distribution of macros and whitespace compared to (b-d), ensuring sufficient spatial resources for downstream standard cell placement (see Appendix~\ref{app:results} for additional visualizations of our method across all benchmark designs).}
\label{fig:placement}
\end{center}
\end{figure*}

\subsection{Experimental Setup}
Our software and hardware environments are as follows: \textsc{Ubuntu} 18.04 LTS, \textsc{Python} 3.8.10, \textsc{Pytorch} 1.8.1 and AMD \textsc{EPYC} 7513 CPU, and \textsc{NVIDIA RTX A6000}. The mean and variance of 10 runs are reported for model evaluation. Hyperparameter settings are detailed in Appendix~\ref{app:hyper}.

\subsubsection{Benchmarks and Dataset Construction}
We conduct our experiments on the Modern Mixed-Size (MMS) benchmarks~\cite{yan2009handling}, modified from ISPD2005~\cite{nam2005ispd2005} with movable macros and fixed I/O pads whose dimensions are set to zero.
Detailed statistics for each benchmark circuit are summarized in Appendix~\ref{app:stat}.

To effectively train our diffusion model to capture generalized topological features, we constructed a dedicated training dataset via data augmentation.
Since the original benchmarks provide limited samples, we generated 1,000 augmented netlists for each of the eight ISPD2005 designs, resulting in a total dataset of 8,000 samples.
The augmentation was performed using a degree-preserving configuration model~\cite{newman2003structure}. 
In this process, the degrees of macro and net nodes are preserved to maintain the fundamental circuit complexity, while edges are randomly rewired to create diverse connectivity patterns.
For each augmented netlist, a high-quality reference macro placement was generated using DREAMPlace to serve as the ground truth ($x_0$) for training.

\subsubsection{Evaluation Protocol and Metrics} \label{sec:metrics}
To assess placement quality, we adopt a unified pipeline across all methods: (i) macro sampling using the benchmark models (e.g., MacroDiff+), (ii) macro legalization, (iii) standard cell placement via DREAMPlace with macros held fixed, and (iv) final legalization. 
The primary metric is the Half-Perimeter Wirelength (HPWL), where lower values indicate better quality.

\subsubsection{Baselines}
We evaluate our framework by comparing it against four representative macro placement methodologies:
\begin{itemize} \item DREAMPlace~\cite{chen2023stronger}: A widely adopted analytical placer serving as the \textit{performance upper bound} for wirelength quality in our benchmarks.
\item MaskPlace~\cite{lai2022maskplace}: A representative RL-based method that formulates placement as a sequential decision process using visual masks. 
\item WireMask-BBO~\cite{shi2023macro}: An optimization-centric approach utilizing wire-mask-guided black-box optimization. 
\item ChipDiffusion~\cite{lee2025chip}: A recent generative method applying diffusion models to chip placement, serving as a direct generative baseline. \end{itemize}

\subsection{Macro Placement Results}
Following the evaluation protocol described in Section~\ref{sec:metrics}, we report the final Half-Perimeter Wirelength (HPWL) after standard cell placement.
Table~\ref{table:results} summarizes the quantitative comparison between MacroDiff+ and state-of-the-art baselines across all benchmark designs.

\subsubsection{Superior Quality and Scalability} 
Across all benchmarks, MacroDiff+ consistently outperforms existing learning-based and optimization-based baselines. 
Compared to the RL-based MaskPlace and the optimization-based WireMask-BBO, our framework achieves substantial HPWL reductions of 6.1\% and 6.2\% on average, respectively. 
More importantly, our method demonstrates superior scalability. 
As shown in Table \ref{table:results}, both MaskPlace and WireMask-BBO failed to converge (T/O) on large-scale designs such as \textit{bigblue2} and \textit{bigblue4} due to computationally prohibitive sequential rollouts and heuristic exploration.
In contrast, MacroDiff+ successfully scales to these massive netlists with superior efficiency. 
Specifically, while the generative baseline ChipDiffusion demonstrates competitive placement quality, it incurs substantial computational overhead on large designs. 
MacroDiff+ alleviates this bottleneck and maintains tractable runtime across all benchmarks (see Appendix~\ref{app:runtime} for detailed runtime analysis).

\subsubsection{Stability and Topology Awareness} 
Compared to ChipDiffusion, a state-of-the-art generative baseline, our method reduces HPWL by 1.2\% on average. 
While this margin might appear modest, a closer inspection reveals a significant advantage in stability. 
For instance, in \textit{adaptec2}, our method exhibits a standard deviation of 2.6, whereas ChipDiffusion shows a high variance of 20.9. 
This indicates that ChipDiffusion's reliance on geometric patterns alone leads to unstable convergence. 
As observed from the generated samples (cf. Fig.~\ref{fig:placement}), MacroDiff+ produces a remarkably uniform distribution of macros across the canvas. 
Unlike RL-based or optimization-centric baselines that often suffer from severe macro clumping—creating massive routing blockages in localized regions—our method distributes whitespace evenly. 
This balanced layout is critical as it secures the necessary spatial slack for the subsequent placement of millions of standard cells, effectively providing favorable conditions for downstream routing.
By explicitly incorporating circuit connectivity via our dual-branch architecture, ours ensures that generated layouts are not only valid but also consistently optimized for global wirelength.

\subsubsection{Closing the Gap with Analytical Solvers} 
Finally, our results significantly narrow the gap between learning-based methods and the analytical upper bound set by DREAMPlace. 
MacroDiff+ achieves placement quality within a gap of 3.2\%--7.3\% relative to DREAMPlace on most benchmarks.
This shows that a physics-guided geometric diffusion model can approach the wirelength quality of traditional analytical optimization while offering the generative diversity inherent to data-driven approaches.

\begin{table}[t]
\centering
\footnotesize
\setlength{\tabcolsep}{1.mm} 
\begin{tabularx}{\columnwidth}{X c c r@{\,$\pm$\,}l c}
\toprule
\multirow{2.5}{*}{Config.} & \multicolumn{2}{c}{Displacement} & \multicolumn{3}{c}{HPWL} \\
\cmidrule(lr){2-3} \cmidrule(l){4-6}
 & Value $\downarrow$ & Degr. (\%) & \multicolumn{2}{c}{Value $\downarrow$} & Degr. (\%) \\
\midrule
MacroDiff+       & 8.7  & -     & 299.3 & 4.3  & - \\
w/o $\epsilon_{net}$  & 11.8 & 36.3  & 297.7 & 8.0  & -0.5 \\
w/o Hetero GNN   & 18.3 & 111.7 & 313.7 & 12.0 & 4.8 \\
w/o Transformer  & 17.0 & 95.4  & 310.5 & 21.3 & 3.7 \\
\bottomrule
\end{tabularx}
\caption{Ablation Study on Dual-Branch Architecture. Comparison of Macro Legalization Results on Displacement ($\times 10^3$) and Mixed-Size Placement Results on HPWL ($\times 10^5$) for \textit{bigblue3}.}
\label{table:architect}
\end{table}

\subsection{Ablation Studies}
\label{sec:ablation}
To validate the effectiveness of each component in MacroDiff+, we conducted extensive ablation studies.
We analyze the impact of our dual-branch architecture and the physics-guided gradient mechanism separately.

\subsubsection{Effectiveness of Dual-Branch Architecture}
To assess the contribution of each architectural component, we compare the full MacroDiff+ framework against three ablated variants:
(1) \textbf{w/o Net-Centric Noise ($\epsilon_{net}$)}, %where the net-centric noise is disabled; 
(2) \textbf{w/o Hetero GNN}, %where the graph module is removed; and
(3) \textbf{w/o Transformer}, %where the transformer module is excluded, relying solely on local GNN features.
where each variant disables the corresponding module.

Table~\ref{table:architect} summarizes the legalization displacement and post-legalization HPWL results. 
The \textbf{MacroDiff+} achieves the best overall balance between legality, wirelength, and stability. 
Interestingly, the \textbf{w/o Net-Centric Noise} configuration results in a slight HPWL decrease (-0.5\%) but incurs a 36.3\% increase in displacement. This suggests that without net-driven gradients, the model over-clusters macros to minimize local wirelength, ignoring global density constraints.

On the other hand, removing the geometric branch (\textbf{w/o Transformer}) results in severe instability. 
This configuration exhibits the highest variance in HPWL (Std: 21.3), nearly $5\times$ that of the full model, and a 95.4\% increase in displacement.
This implies that while the GNN captures connectivity, it fails to construct a valid global floorplan, leading to massive overlaps that require drastic legalization.
Finally, the absence of the Hetero GNN (\textbf{w/o Hetero GNN}) leads to the worst wirelength degradation (+4.8\%) and displacement (+111.7\%), confirming that the macro--net distinction is the most critical factor for topology-aware sampling.

\begin{table}[t]
\centering
\footnotesize 
\setlength{\tabcolsep}{1.mm}
\begin{tabularx}{\columnwidth}{X c c r@{\,$\pm$\,}l c}
\toprule
\multirow{2.5}{*}{Config.} & \multicolumn{2}{c}{Displacement} & \multicolumn{3}{c}{HPWL} \\
\cmidrule(lr){2-3} \cmidrule(l){4-6}
 & Value $\downarrow$ & Degr. (\%) & \multicolumn{2}{c}{Value $\downarrow$} & Degr. (\%) \\
\midrule
MacroDiff+       & 8.7   & -      & 299.3 & 4.3  & - \\
Overlap Guidance & 18.5  & 114.0  & 314.1 & 28.7 & 5.0 \\
No Guidance      & 108.1 & 1147.4 & 305.7 & 8.0  & 2.1 \\
\bottomrule
\end{tabularx}
\caption{Ablation Study on Gradient-Guided Sampling. Comparison of Macro Legalization Results on Displacement ($\times 10^3$) and Mixed-Size Placement Results on HPWL ($\times 10^5$) for \textit{bigblue3}.}
\label{table:guidance}
\end{table}

\subsubsection{Impact of Gradient Guidance}
To evaluate the impact of our explicit physical steering, we analyze the loss components used in the sampling process (Eq.~\ref{eq:guide}). 
We compare three strategies: 
(1) \textbf{MacroDiff+} (Full Guidance), applying both wirelength (HPWL) and overlap penalties; 
(2) \textbf{Overlap Guidance}, which focuses solely on resolving physical intersections; and 
(3) \textbf{No Guidance}, where the diffusion process proceeds without any gradient injection.

Table~\ref{table:guidance} reports the results. 
\textbf{MacroDiff+} yields superior placement quality, achieving both the lowest legalization displacement and downstream HPWL. 
In the \textbf{No Guidance} setting, the trajectory becomes physically unconstrained. 
While the HPWL degradation (+2.1\%) appears moderate, the displacement explodes by over $11\times$ (+1147.4\%). 
This indicates that without physical forces, the model blindly clusters macros to minimize wirelength, disregarding feasibility, which forces the legalizer to destroy the layout structure.

Conversely, \textbf{Overlap Guidance} reduces displacement compared to the unguided baseline but degrades HPWL by 5.0\% and exhibits the highest variance (Std: 28.7). 
This suggests that focusing solely on legality causes macros to \textit{over-separate} to avoid overlaps, breaking critical net connections. 
Thus, the joint physics guidance is essential: the HPWL gradient pulls connected macros together, while the overlap gradient ensures they remain valid, producing placements that are simultaneously legal and connectivity-efficient.

\section{Conclusion}
In this paper, we presented MacroDiff+, a physics-guided geometric diffusion framework that addresses the fundamental challenges of macro placement in modern VLSI design. 
Our dual-branch architecture synergizes topological insights from Heterogeneous GNNs with global geometric contexts from Transformers to capture the complex joint distribution of circuit connectivity and spatial constraints. 
Furthermore, the Physics-Guided Sampling strategy actively steers the generative trajectory using explicit physical gradients, ensuring that solutions are both statistically plausible and physically robust. 
Notably, the uniform distribution of whitespace in our placements leaves sufficient spatial slack for standard cell placement, as reflected in the consistent HPWL gains after the full mixed-size placement flow.
Experimental results on ISPD2005 MMS benchmarks demonstrate that our approach outperforms existing baselines in stability and scalability, narrowing the gap between generative placement methods and analytical solvers. 
%Future work includes validation on modern designs to assess cross-family generalization.
We believe this work takes a step toward topology-aware generative placement in physical design.

%\newpage
\section*{Acknowledgments}
This research was supported by the Nano \& Material Technology Development Program through the National Research Foundation of Korea (NRF) funded by Ministry of Science and ICT (RS-2022-NR068233), the National Research Foundation of Korea (NRF) grant funded by the Korea government (MSIT) (RS-2024-00405991), and by the POSTECH Institute of Artificial Intelligence.

\section*{Contribution Statement}
Jongho Yoon and Jinsung Jeon contributed equally to this work as co-first authors. Jinsung Jeon conducted this research while at the University of California, San Diego.

% %% The file named.bst is a bibliography style file for BibTeX 0.99c
\bibliographystyle{named}
\bibliography{ijcai26}

\clearpage
\appendix

\section{Preliminaries on Diffusion Models}
\label{app:diffusion}
In this section, we provide the detailed mathematical formulation of the Denoising Diffusion Probabilistic Models (DDPMs) used in our framework.

\subsection{Forward Process}
The forward process is a fixed Markov chain that incrementally adds noise according to a variance schedule $\beta_t$:
\begin{equation}
    q(x_t | x_{t-1}) := \mathcal{N}(x_t; \sqrt{\alpha_t} x_{t-1}, \beta_t \mathbf{I}), 
\end{equation}
where $\alpha_t = 1 - \beta_t$. A key property allows direct sampling of $x_t$ from $x_0$:
\begin{equation} 
\label{eq:forward}
    x_t = \sqrt{\bar{\alpha}_t} x_0 + \sqrt{1 - \bar{\alpha}_t} \epsilon, \quad \epsilon \sim \mathcal{N}(0, \mathbf{I}),
\end{equation}
where $\bar{\alpha}_t = \prod_{s=1}^t \alpha_s$.

\subsection{Reverse Process}
The reverse process approximates the intractable posterior $q(x_{t-1}|x_t)$ using a learned Gaussian transition:
\begin{equation}
    p_\theta(x_{t-1} | x_t) = \mathcal{N}(x_{t-1}; \mu_\theta(x_t, t), \Sigma_\theta(x_t, t)). 
\end{equation}
The model is trained to predict the mean $\mu_\theta$ (parameterized via noise prediction $\epsilon_\theta$) to effectively denoise the sample.

\section{Detailed Input Features}
\label{app:features}
Table~\ref{table:feats_resized} provides the detailed specifications of the node and edge features used in our heterogeneous graph (Section~\ref{sec:formulation}).
To effectively guide the diffusion model in generating physically valid and topologically optimized placements, we distinguish between \textit{static features} (which describe invariant circuit properties) and \textit{dynamic features} (which evolve during the denoising process).

\textbf{Static Features.} The macro size ($w, h$), chip area boundaries, and net degrees serve as the fundamental constraints. Notably, the \textit{Pin Offset} feature on edges is critical for large-scale macros, as it allows the model to calculate wirelength using exact pin locations rather than node centers.

\textbf{Dynamic Features.} The \textit{Position} ($x_t$) and \textit{HPWL} are updated at every diffusion step. By dynamically embedding the current HPWL into net nodes, the model receives real-time feedback on congestion and routing quality, enabling gradient-guided refinement.

\begin{table}[b]
\centering
\vspace{-0.1cm}
\renewcommand{\arraystretch}{0.83}
\renewcommand{\tabcolsep}{0.7mm}
\scriptsize
\begin{tabularx}{\columnwidth}{c c X c}
\toprule
\textbf{Type} & \textbf{Feature} & \multicolumn{1}{c}{\textbf{Description}} & \textbf{Dim.} \\
\midrule
Global & \textit{Chip Area} & Defines the absolute boundaries (width and height) of the core region providing global spatial context. & 2 \\
\midrule
\multirow{2.5}{*}{\begin{tabular}[c]{@{}c@{}}Macro Node\\ ($\mathcal{V}_m$)\end{tabular}} 
& \textit{Size} & The physical dimensions ($w, h$) of each macro. Critical constraint for avoiding overlaps. & 2 \\
\cmidrule(lr){2-4}
& \textit{Position} & The 2D coordinates ($p_x, p_y$). Represents the noisy state $x_t$ and is the target for denoising. & 2 \\
\midrule
\multirow{2.5}{*}{\begin{tabular}[c]{@{}c@{}}Net Node\\ ($\mathcal{V}_n$)\end{tabular}} 
& \textit{HPWL} & Half-Perimeter Wirelength. A proxy for congestion, calculated from noisy positions $x_t$. & 1 \\
\cmidrule(lr){2-4}
& \textit{Degree} & Total count of connected macros and standard cells. Serves as a static connectivity indicator. & 2 \\
\midrule
Edge ($\mathcal{E}$) & \textit{Pin Offset} & Relative 2D displacement ($o_x, o_y$) of a pin from the macro center for accurate modeling. & 2 \\
\bottomrule
\end{tabularx}
\vspace{-0.1cm}
\caption{Summary of Input Features for the Circuit Graph.}
\label{table:feats_resized}
\end{table}

\section{Hyperparameter Settings} \label{app:hyper}
Table~\ref{table:hyper} summarizes the hyperparameters used to ensure reproducibility, including the maximum guidance steps (\texttt{ITER}, $K$), the learning rate (\texttt{LR}, $eta$), and the HPWL plateau threshold (\texttt{TH}) for transitioning between Phase 1 and 2. These optimal configurations vary across designs because the magnitudes of $L_{HPWL}$ and $L_{overlap}$ are highly sensitive to design-specific properties, such as chip dimensions and macro count.

\begin{table}[h]
\centering
%\vspace{-0.1cm}
\renewcommand{\arraystretch}{0.9}
\renewcommand{\tabcolsep}{1.5mm}
{\footnotesize
\begin{tabular}{l c c c l c c c}
\toprule
%Designs & \texttt{ITER} & \texttt{LR} & \texttt{TH} & \texttt{WT} \\
Designs & \texttt{ITER} & \texttt{LR} & \texttt{TH} & Designs & \texttt{ITER} & \texttt{LR} & \texttt{TH} \\
\cmidrule(r){1-4} \cmidrule(){5-8}
\textit{adaptec1} & 300 & 0.05 & 0.5 & \textit{bigblue1} & 300 & 0.05 & 0.1 \\
\textit{adaptec2} & 700 & 0.05 & 0.5 & \textit{bigblue2} & 700 & 0.01 & 0.05 \\
\textit{adaptec3} & 700 & 0.05 & 0.1 & \textit{bigblue3} & 700 & 0.005 & 0.1 \\
\textit{adaptec4} & 700 & 0.05 & 0.1 & \textit{bigblue4} & 500 & 0.05 & 0.05 \\
\bottomrule
\end{tabular}
}
\vspace{-0.1cm}
\caption{The best hyperparameters}
\vspace{-0.3cm}
\label{table:hyper}
\end{table}

\section{Benchmark Statistics}
\label{app:stat}

Table~\ref{table:stat} summarizes the detailed statistics of the ISPD2005 MMS benchmarks used in our experiments.
To focus on the macro placement problem, we construct the circuit graph by retaining only the components relevant to macro connectivity.
Specifically, \# Net and \# Pin indicate the number of nets and pins, respectively, that are connected to at least one macro or fixed I/O pad.
The diversity in the number of macros and connectivity complexity across these designs ensures robust evaluation of scalability and generalization.

\begin{table}[h]
\centering
%\vspace{-0.1cm}
\renewcommand{\arraystretch}{0.9}
{\footnotesize
\begin{tabular}{ccccc}
\toprule
Designs  & \# Macros & \# I/O & \# Net & \# Pin \\
\midrule
\textit{adaptec1} & 63       & 480   & 10552 & 17606 \\
\textit{adaptec2} & 127      & 439   & 12386 & 20965 \\
\textit{adaptec3} & 58       & 665   & 9854  & 15648 \\
\textit{adaptec4} & 69       & 1260  & 13069 & 19587 \\
\textit{bigblue1} & 32       & 528   & 6727  & 9334  \\
\textit{bigblue2} & 959      & 22125 & 4001  & 10918 \\
\textit{bigblue3} & 69       & 1229  & 17537 & 26085 \\
\textit{bigblue4} & 199      & 7971  & 39637 & 69952 \\
\bottomrule
\end{tabular}
}
\vspace{-0.1cm}
\caption{Statistics of the ISPD2005 modern mixed-size benchmarks.} %Only nets and pins connected to at least one macro are retained.}
\vspace{-0.3cm}
\label{table:stat}
\end{table}

\section{Runtime Analysis} \label{app:runtime}

As shown in Table~\ref{table:runtime}, MacroDiff+ scales gracefully with circuit size (30–91s), whereas the overhead of ChipDiffusion on bigblue2 and bigblue4 stems primarily from its built-in legalization procedure, which scales poorly with macro count.

\begin{table}[b]
\centering
%\vspace{-0.1cm}
\renewcommand{\arraystretch}{0.9}
{\footnotesize
\begin{tabular}{l c c c}
\toprule
%Designs & \texttt{ITER} & \texttt{LR} & \texttt{TH} & \texttt{WT} \\
Designs & DREAMPlace & ChipDiffusion & Ours \\
\midrule
\textit{adaptec1} & 9.6 & 142.2 & 30.4 \\
\textit{adaptec2} & 18.3 & 141.5 & 70.6 \\
\textit{adaptec3} & 23.8 & 145.1 & 71.4 \\
\textit{adaptec4} & 19.8 & 170.2 & 71.5 \\
\textit{bigblue1} & 11.2 & 140.4 & 30.9 \\
\textit{bigblue2} & 20.2 & 17274.2 & 73.6 \\
\textit{bigblue3} & 89.2 & 169.1 & 70.8 \\
\textit{bigblue4} & 113.2 & 2427.4 & 90.9 \\
\bottomrule
\end{tabular}
}
\vspace{-0.1cm}
\caption{Comparison of inference runtime (in seconds) for macro placement generation.}
%\vspace{-0.1cm}
\label{table:runtime}
\end{table}

\section{Qualitative Placement Results} \label{app:results}

\begin{figure}[!ht]
\centering
\includegraphics[width=85.1mm]{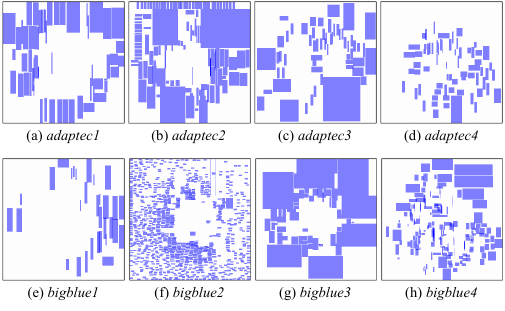}
\vspace{-0.7cm}
\caption{Visualization of macro placement generated by MacroDiff+ on the ISPD2005 MMS benchmark suite.}
\label{app:resultsofours}
\vspace{-0.4cm}
\end{figure}

Fig.~\ref{app:resultsofours} visualizes macro placements generated by MacroDiff+ across all ISPD2005 MMS benchmark designs.

\end{document}